\documentclass[10pt,conference,compsocconf]{IEEEtran}

\usepackage{hyperref}
\usepackage{graphicx}	
\usepackage[ruled,vlined]{algorithm2e}
\usepackage{amsmath,amsfonts,amssymb,amsthm}
\usepackage{amssymb}
\usepackage{mathtools}
\usepackage{commath}
\usepackage{siunitx}
\usepackage{subfig}

\begin{document}
\title{Federated Learning with Differential Privacy}

\author{
    \IEEEauthorblockN{Adrien Banse\IEEEauthorrefmark{1}, Jan Kreischer\IEEEauthorrefmark{2}, Xavier Oliva i Jürgens\IEEEauthorrefmark{3}}
    \IEEEauthorblockA{\IEEEauthorrefmark{1}Exchange student at EPFL, Switzerland from UCLouvain, Belgium}
    \IEEEauthorblockA{\IEEEauthorrefmark{2}Exchange student at EPFL, Switzerland from Universität Zürich, Switzerland}
    \IEEEauthorblockA{\IEEEauthorrefmark{3}Student at EPFL, Switzerland}
}

\maketitle

\begin{abstract}

Federated learning (FL), as a type of distributed machine learning, is capable of significantly preserving client's private data from being shared among different parties. Nevertheless, private information can still be divulged by analyzing uploaded parameter weights from clients. In this report, we showcase our empirical benchmark of the effect of the number of clients and the addition of differential privacy (DP) mechanisms on the performance of the model on different types of data. Our results show that non-i.i.d and small datasets have the highest decrease in performance in a distributed and differentially private setting.

\end{abstract}

\section{Introduction}
\label{intro}

The training of deep-learning models usually requires large and diverse datasets. In many areas, gathering large datasets is difficult and requires the collaboration of multiple institutions. Especially in medicine, patient data is spread among multiple entities. While each party might not have enough data to robustly train a model for a specific task, the union of their datasets can potentially lead to successful data insights. Especially for rare diseases, data sharing among many entities, which can be located in different countries, is crucial. However, medical data contains much private information and is potentially identifiable, making it especially sensitive with regard to privacy. Legal regulations, such as GDPR in Europe, contain specific clauses describing the privacy requirements that medical data has to comply with.

To address these important limitations, privacy-preserving techniques are gaining momentum, as they can enable to perform machine learning (ML) algorithms on sensitive data. Especially federated machine learning, a non-cryptographic approach for privacy-preserving training of ML models, has been increasingly studied in the last years \cite{mcmahan2017communicationefficient, geyer2018differentially, tramer2021differentially}. This will be covered in section \ref{fed}.

Another privacy-enhancing technology used for training ML models is differential privacy (DP). DP algorithms aim at quantifying and setting an upper-bound to the privacy loss of an individual when entering their private data into a dataset. They rely on incorporating random noise to the data or model. DP has also been used in the federated setting \cite{wei2019federated}, where to collectively train a model, multiple parties exchange or send differentially private model updates to a central server to protect against an honest-but-curious adversary. This will be covered in section \ref{dp}.

In this project, we will focus on cross-silo federated machine learning with differential privacy.
The three questions we want to answer are the following.

1) How does the level of data distribution affect model accuracy, i.e. the convergence of the optimization process?

2) How does differential privacy affect model accuracy?

3) Is it applicable to small, more realistic datasets?

We will perform experiments for both i.i.d., where all of the data is independently and identically distributed, and non-i.i.d cases. Finally, we will apply our FL-DP algorithm on a small medical dataset.

\section{Theoretical Background}

\label{fed}
\subsection{Federated Machine Learning}

In federated learning, the data remains under the control of their owners, which we designate as clients, and a central server coordinates the training by sending the global model directly to the clients, which then update the model with their data. The updated models from the clients are sent back to the central server and averaged to obtain an updated version of the global model. This is done via the \texttt{FedAvg} (Federated Averaging) algorithm \cite{mcmahan2017communicationefficient}, described in Algorithm \ref{fedavg} in Appendix~\ref{appendix}.

\subsection{Differential Privacy}
\label{dp}

In this project, we focus on cross-silo federated learning, where data is distributed between organizations with high computational resources, like hospitals or banks. The biggest challenge in this setting usually lies on the data security side.

The adversarial model in this setting includes both the central server and any of the clients. The central server observes the updated parameters of all clients, while a client observes its own updates and the new global model parameters after every round. The problem lies in the fact that the model parameters might leak information about the training data. The adversary’s goal is to infer whether a given record was in the client’s training data (membership inference) or learn properties about the client's training data (property inference).

($\epsilon$, $\delta$)-DP provides a strong criterion for privacy preservation of distributed data processing systems. Here, $\epsilon > 0$ is the distinguishable bound of all outputs on two neighboring datasets (pairs of databases $(\mathcal{D}, \mathcal{D}_{-r})$ differing only in one row $r$. In other words, the removal or addition of a single record in the database should not substantially affect the values of the computed function/statistics.
$$\log{\frac{\text{P}[A(\mathcal{D}) = O]}{\text{P}[A(\mathcal{D}_{-r}) = O]}} < \epsilon \text{ with probability } 1 - \delta$$
Thus, $\delta$ represents the probability that the ratio of the probabilities for two neighboring datasets cannot be bounded by $e^\epsilon$). Typically, values of $\delta$ that are less than the inverse of any polynomial in the size of the database are used.

There are three ways to apply differential privacy to Machine Learning: objective perturbation, gradient perturbation and output perturbation.
For deep learning applications, we can not derive sensitivity bounds for the objective and output and have to use gradient perturbation.
We use \texttt{PyTorch}'s module \texttt{Opacus}, that uses gradient perturbation with the advanced composition method. It injects noise at every iteration by using the gradient clipping technique \cite{Abadi_2016} (see Algorithm~\ref{clientupdate} in Appendix~\ref{appendix}).


\section{Models and Methods}

\subsection{Datasets and models}
\label{data}

In this section, we shortly describe the datasets used for the numerical experiments, as well as the model used to classify them.

\subsubsection{MNIST}
is a widely used database comprising thousands of $28 \times 28$ pixels images of 10 different handwritten digits (which means there are 10 different classes). The samples are randomly and equally distributed among clients, we say that the data is i.i.d.. We use the Convolution Neural Network (CNN) defined in the \texttt{PyTorch}'s examples GitHub repository \cite{mnist_pytorch_example}. 60 000 data points were used for training, and 10 000 for testing.

\subsubsection{FEMNIST}
(from the LEAF benchmark \cite{caldas2019leaf}) is also a database consisting of $28 \times 28$ images of 10 different handwritten digits (letters are discarded). The difference between MNIST and FEMNIST datasets is that the partitioning of the data is now based on the writer of the digit. It means that the underlying distribution of data for each user is now consistent with the raw data, yielding a non-i.i.d. dataset. We use the same CNN, and same training/testing dataset sizes as for MNIST.

\subsubsection{Medical dataset}
In order to tackle more realistic scenarios as explained in Section~\ref{intro}, we use a database of medical records of 120 patients \cite{med_dataset}. The aim is to predict whether patients suffer from inflammation of urinary bladder and/or nephritis of renal pelvis origin. In this work, we focus on the pathology. The medical data is randomly split onto the clients, representing hospitals. There are 6 attributes, namely temperature of the patient, occurrence of nausea, lumbar pain, urine pushing, micturition pains, and burning of urethra, itch, swelling of urethra outlet. We used logistic regression to tackle this classification problem. 96 data points were used for training, and 24 for testing.
 
\subsection{Experimental setup}
The parameters for the experiments are set as follows:
The learning rate of SGD is set to $0.01$. The number of client epochs per round is set to 1 (to avoid the clients falling in different local minima) for the MNIST and FEMNIST database, 100 for the medical dataset.
The number of global rounds is set to 30. We set the batch size of all clients to 128 for MNIST and FEMNIST, and 8 for the medical dataset.

\subsubsection{First Experiment}

In order to see the impact of federated learning, we will leave all training parameters fixed and perform federated learning for $\texttt{nr\_clients} \in \{1, 5, 10\}$ Note that the amount of data points remains constant but more distributed.

\subsubsection{Second Experiment}

We reuse the previous training setting, but fix $\texttt{nr\_clients} = 10$.
Now, we want to analyze the effect of differential privacy on performance and convergence. 
We set the clipping threshold of the gradient to $1.2$.
We set the privacy parameter $\delta$ to $\frac{1}{2n}$.
We compare the training loss and final accuracy with various protection levels for 10 clients, using $\epsilon \in \{10, 50, 100\}$. Every round, every client uses a privacy budget of $\frac{\epsilon}{\texttt{nr\_rounds}}$.

We report the testing accuracy of the global model, as well as its loss for every training round.

\section{Results}
\label{results}

\subsection{MNIST}
\label{expmnist}

\begin{figure}[htp] 
    \centering
    \subfloat[Accuracy and loss for different numbers of clients.]{%
        \includegraphics[width=0.48\columnwidth]{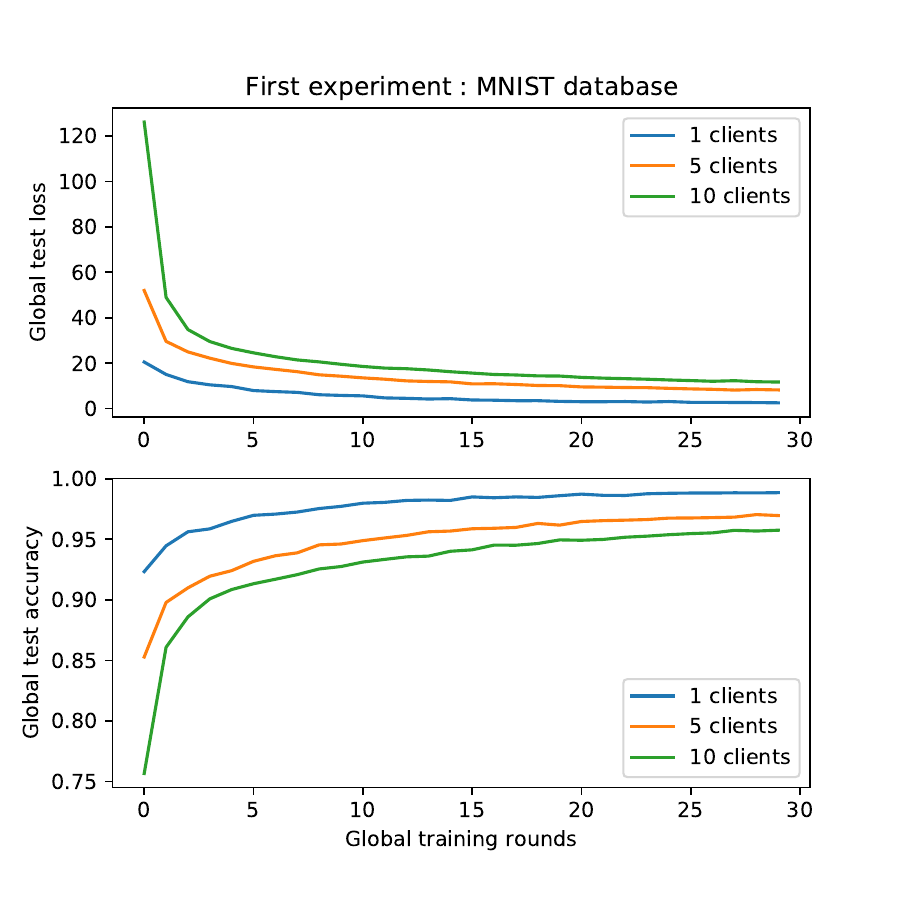}%
        \label{mnist_exp1}%
        }%
    \hfill%
    \subfloat[Accuracy and loss for different values of $\epsilon$, with 10 clients.]{%
        \includegraphics[width=0.48\columnwidth]{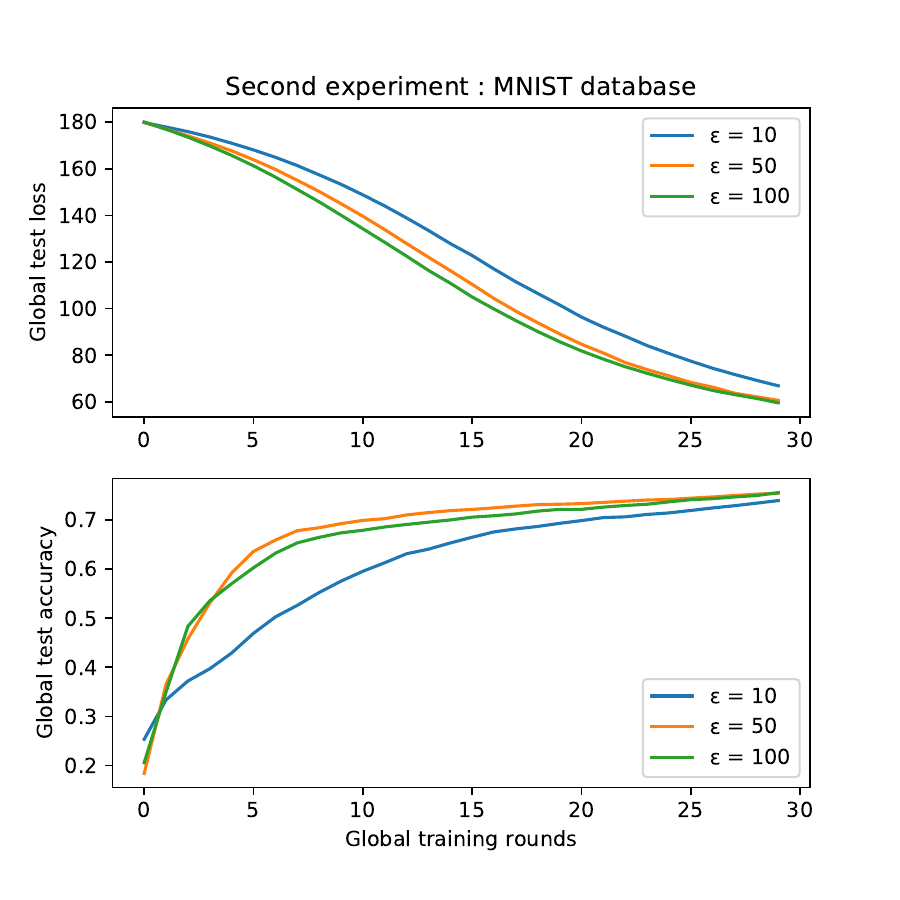}%
        \label{mnist_exp2}%
        }
        \caption{MNIST database}
\end{figure}

Results of the first experiment are shown in Figure \ref{mnist_exp1}, and results of the second experiment are shown in Figure \ref{mnist_exp2}. 

Concerning the first experiment, as expected, one can clearly observe that the more clients there are, the slower the learning process. 

Regarding the second experiment, one can see that the learning is faster when we go from a privacy budget of $\epsilon=10$ to $\epsilon=50$. The same observation can be made for the transition from $\epsilon=50$ to $\epsilon=100$ by looking at the testing loss graph. However, we can observe that the improvement is not proportional to $\epsilon$ since we observe a less important change for the second transition. The three privacy setups lead to the same final accuracy though. 

Finally, one can observe on Figure \ref{mnist_exp1} and Figure \ref{mnist_exp2} the difference of convergence of SGD with and without privacy in a federated learning context. Without privacy, SGD achieves more than $\SI{95}{\percent}$ of testing accuracy, while only $\SI{75}{\percent}$ with privacy, even with a large privacy budget such as $\epsilon = 100$.

\subsection{FEMNIST}

\begin{figure}[htp!] 
    \centering
    \subfloat[Accuracy and loss for different numbers of clients for the FEMNIST database.]{%
        \includegraphics[width=0.48\columnwidth]{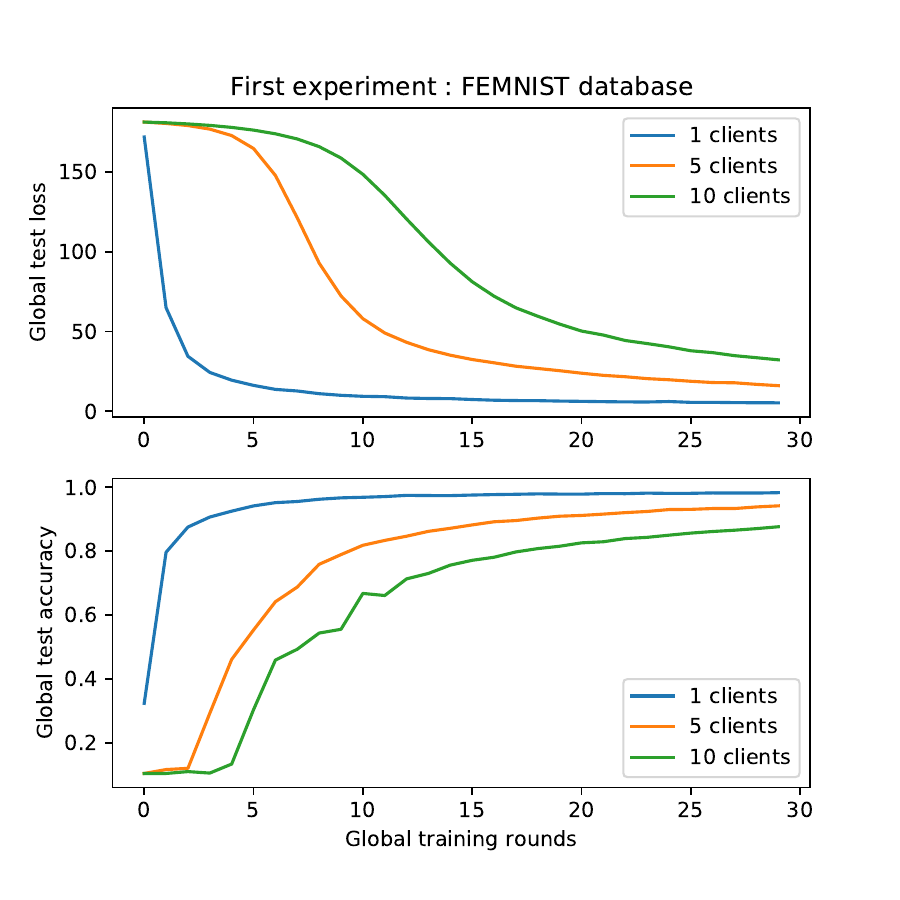}%
        \label{femnist_exp1}%
        }%
    \hfill%
    \subfloat[Accuracy and loss for different values of $\epsilon$, with 10 clients for the FEMNIST database.]{%
        \includegraphics[width=0.48\columnwidth]{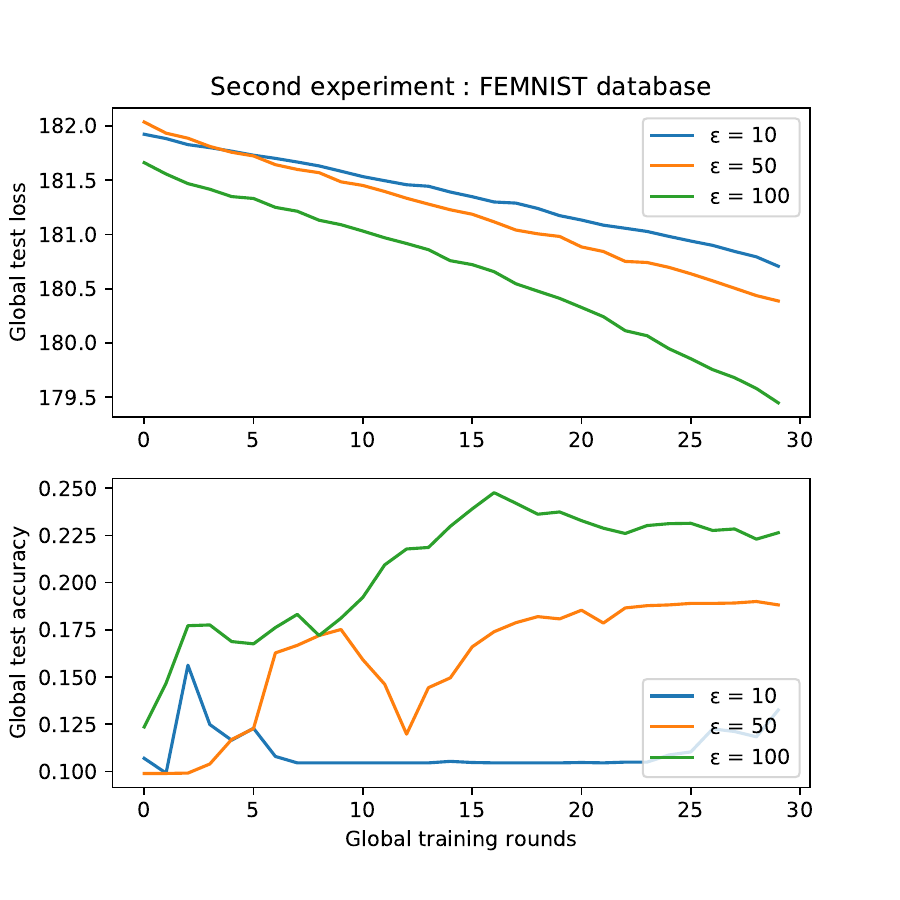}%
        \label{femnist_exp2}%
        }%
    \caption{FEMNIST database}
\end{figure}

Results can be founds in Figure~\ref{femnist_exp1} and Figure~\ref{femnist_exp2}.

We enlighten two main differences compared to Section~\ref{expmnist}:

1) While a single client achieves similar accuracy to the regular MNIST dataset, the algorithm takes more global training rounds to converge with an increasing number of clients. After 100 rounds the test accuracy of all models reaches around 98\%.

2) When using DP the training process does not really converge anymore, not leading to a viable model for FEMNIST.

\subsection{Medical dataset}

\begin{figure}[htp] 
    \centering
    \subfloat[Accuracy and loss for different numbers of clients for the medical database.]{%
        \includegraphics[width=0.48\columnwidth]{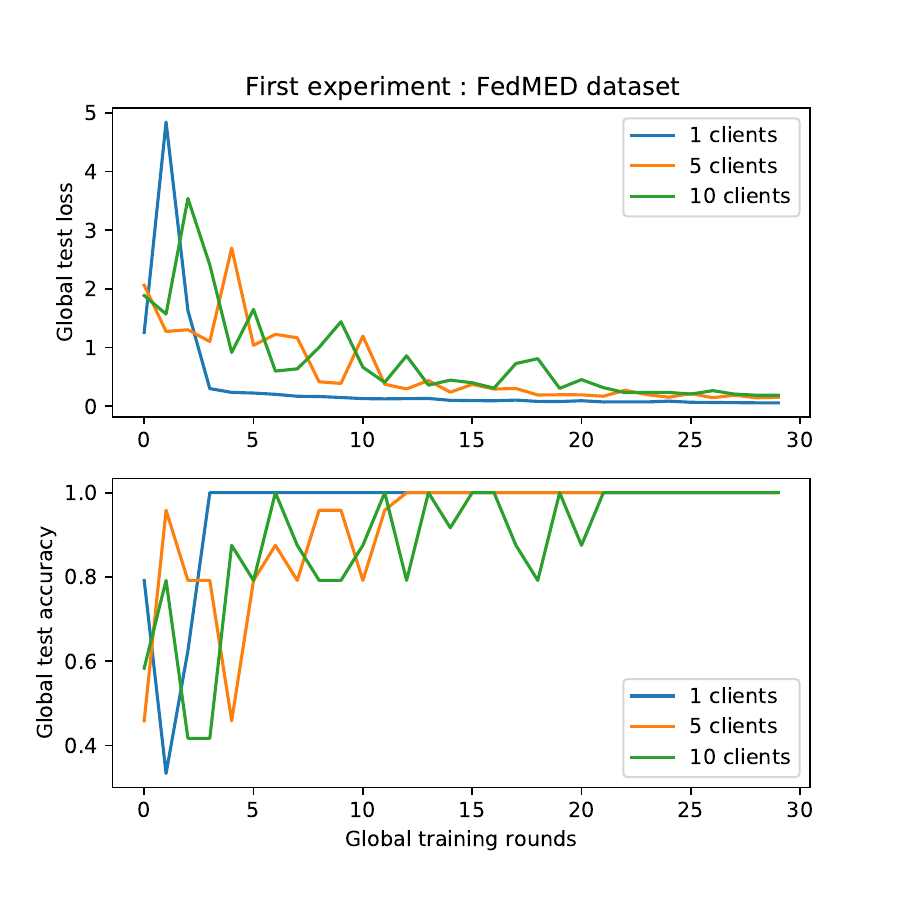}%
        \label{fedmed_exp1}%
        }%
    \hfill%
    \subfloat[Accuracy and loss for different values of $\epsilon$, with 10 clients for the medical database.]{%
        \includegraphics[width=0.48\columnwidth]{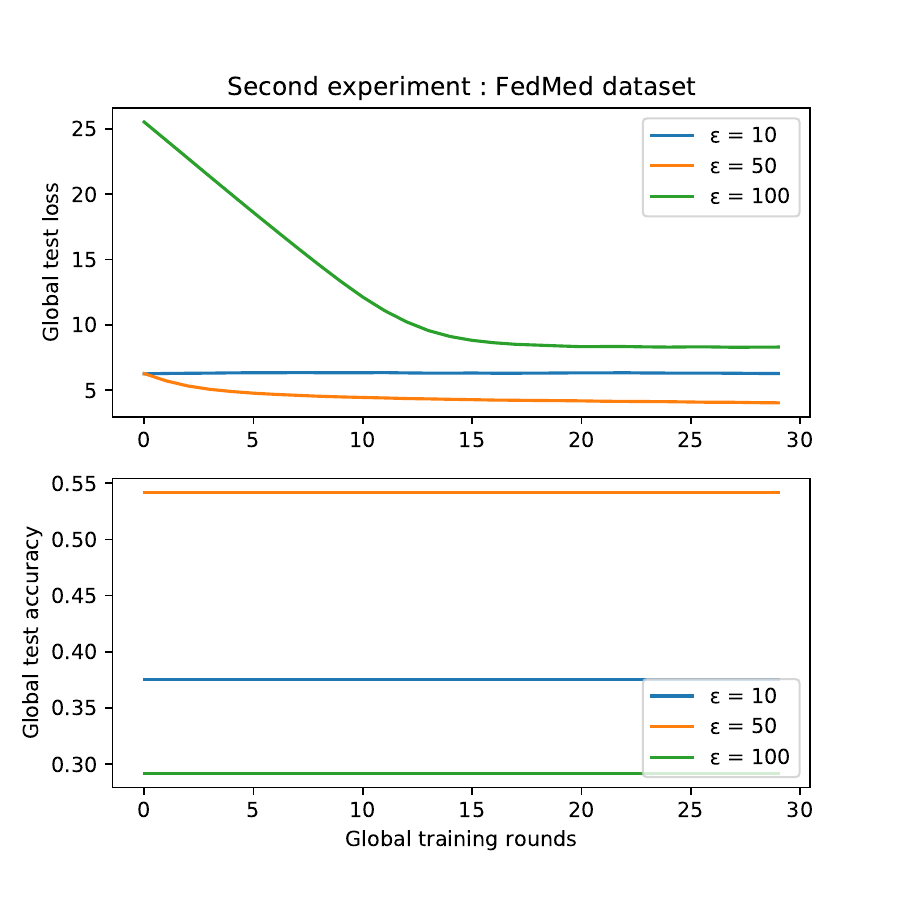}%
        \label{fedmed_exp2}%
        }%
    \caption{Medical database}
\end{figure}

Results can be found on Figure~\ref{fedmed_exp1} and Figure~\ref{fedmed_exp2}.

Again, we enlighten some differences compared to Section~\ref{expmnist}.

1) Training shows higher variance with more clients, but all manage to converge.

2) Adding differential privacy, the model accuracy does not get any better. Only for $\epsilon = 10$ the loss decreases, indicating that the global model is improving.

\section{Discussion}

In this section we will discuss the results in Section~\ref{results} and mention further improvements. 

During the experiments, we clearly observed the trade-off between privacy and utility. Since, participating clients only optimize using their own data, they can end up in different local minima, making the global model sub-optimal. This leads to a generally slower convergence with a higher number of clients.

The experiments for the MNIST dataset show that adding noise and gradient clipping makes the maximal accuracy decrease around $\SI{20}{\percent}$ for the same number of training rounds, even on i.i.d data. More clients might be needed to average out the noisy updates \cite{yu2020salvaging}.

When it comes to non-i.i.d data, distribution of data in silos makes the global model perform worse. A possible improvement would be to use a d-clique topology (decentralized learning) instead of classical centralized federated learning.

Regarding small datasets, i.e. more realistic datasets, the results are more worrying: training is more sensitive to DP mechanisms, making it very difficult to learn while effectively preserving privacy. Additionally, they have a higher variance during training.
We had to use bigger values than the standard $\epsilon = 0.1$, because such a bound was too restrictive to compute a single mini-batch SGD iteration.

When it comes to limitations of our work, we have to note that we did not focus on hyperparameter tuning and used fix parameters that might not be optimal for the task. Additionally, there are other ways like fully homomorphic encryption (FHE) that protects the data from attacks by encrypting the model.

\section{Summary}
We were able to show that models trained using federated machine learning are able to achieve similar accuracy as models trained on a centralized dataset. However, adding DP mechanisms to preserve the privacy of the data showed a rapid decrease of performance, especially for non-i.i.d or small datasets, which are much more realistic datasets. 

\bibliographystyle{IEEEtran}
\bibliography{literature}

\appendix

\subsection{Algorithms}
\label{appendix}

This section groups algorithms used to train global and client models, using differential privacy (see Section~\ref{fed} and Section~\ref{dp}).

\begin{algorithm}[h]
\SetAlgoLined
\KwResult{Global model trained}
\textbf{Initialize: } $\boldsymbol{w^0}$ \;
 \For{\textup{each global round $t=0,1,\dots$}}{
    \For{\textup{each client $k \in \{1, \dots K\}$}}{
        $\boldsymbol{w_k^{t+1}} \gets$ \texttt{ClientUpdate}($k,\boldsymbol{w^t}$)
    }
    $\boldsymbol{w^{t+1}} = \sum_{k=1}^K \frac{n_k}{n} \boldsymbol{w^{t+1}_k}$\;
 }
 \caption{\texttt{FedAvg} (Federated Averaging)}
 \label{fedavg}
\end{algorithm}

In Algorithm \ref{fedavg}, $K$ is the number of clients (each client denoted by $k$), $\alpha$ is the learning rate, $n_k$ is the size of the dataset of client $k$, $n = \sum_{k=1}^K n_k$ is the size of the entire dataset. Moreover, $\boldsymbol{w^t_k} \in \mathbb{R}^{P}$ is the vector of $P$ parameters of the local model of client $k$, at global round $t$, and $\boldsymbol{w^t} \in \mathbb{R}^P$ is the vector of $P$ parameters of the global model. 

The function \texttt{ClientUpdate} involes privacy (see Section~\ref{dp}), and is described in Algorithm~\ref{clientupdate}.

\begin{algorithm}[h]
\SetAlgoLined
\KwResult{$\boldsymbol{w_k^{t+1}}$ and privacy cost ($\epsilon$, $\delta$) using a privacy accounting method}
\textbf{Input:} Parameters of the global model at time $t$ $\boldsymbol{w^t}$\\
 \For{\textup{$\text{epoch } e = 1, ..., E$}}{
 Sample $\lfloor \frac{n}{B} \rfloor$ batches at random\\
     \For{\textup{$j = 1, \dots, \lfloor \frac{n}{B} \rfloor$}}{
     \For{\textup{$i = 1, \dots, B $}}{
    \textbf{Compute gradient}\\
    $\boldsymbol{g_j(x_i^j)} \gets \nabla\mathcal{L}(\boldsymbol{w_k}, \boldsymbol{x_i^j})$\\
    \textbf{Clip gradient}\\
    $\boldsymbol{\Bar{g}_j(x_i^j)} \gets \boldsymbol{g_j(x_i^j)} / \text{max}\left(1, \frac{\norm{\boldsymbol{g_j(x_i^j)}}_2}{C}\right)$\\
    }
    \textbf{Add noise}\\
    $\boldsymbol{\Tilde{g}_j} \gets \frac{1}{B}\left(\sum_{i=0}^B\boldsymbol{\Bar{g}_j(x_i^j)} + \mathcal{N}(0, \sigma^2 C^2 \boldsymbol{I})\right)$\\
    \textbf{Descent}\\
    $\boldsymbol{w_k} \gets \boldsymbol{w_k} - \eta \boldsymbol{\Tilde{g}_j}$
    
    }
 }
 \caption{\texttt{ClientUpdate} for client $k$}
 \label{clientupdate}
\end{algorithm}

In Algorithm \ref{clientupdate}, $\eta$ is the learning rate, $\sigma$ is the noise scale, $C$ is the gradient norm bound, $B$ is the batch-size, $E$ is the total number of epochs, $n$ is the size of the dataset and $\boldsymbol{x}_i^j$ is the $i$-th datapoint in the $j$-th batch. $\mathcal{L}(\boldsymbol{w},\boldsymbol{x})$ is the loss corresponding to the datapoint $\boldsymbol{x}$, depending on the parameters $\boldsymbol{w}$.

\end{document}